\documentclass[10pt,twocolumn,letterpaper]{article}

\usepackage{cvpr}
\usepackage{times}
\usepackage{epsfig}
\usepackage{graphicx}
\usepackage{amsmath}
\usepackage{amssymb}
\usepackage{placeins}
\usepackage{subcaption}
\usepackage{caption}
\usepackage[breaklinks=true,bookmarks=false]{hyperref}

\cvprfinalcopy % *** Uncomment this line for the final submission

\def\cvprPaperID{****} % *** Enter the CVPR Paper ID here

\begin{document}

%%%%%%%%% TITLE
\title{Medical Time Series Classification with Hierarchical Attention-based Temporal Convolutional Networks: A Case Study of Myotonic Dystrophy Diagnosis}
%Interpretable Diagnosis of Myotonic Dystrophy from Handgrip Time Series Data with Hierarchical Attention-based Temporal Convolutional Networks
\author{Lei Lin, Beilei Xu, Wencheng Wu, Trevor Richardson, Edgar A. Bernal\\
Rochester Data Science Consortium\\
1219 Wegmans Hall, University of Rochester\\
250 Hutchison Road, Rochester, NY\\
{\tt\small \{Lei.Lin, Beilei.Xu, Wencheng.Wu, Trevor.Richardson, Edgar.Bernal\}@rochester.edu}
%\and
}

\maketitle
%\thispagestyle{empty}

%%%%%%%%% ABSTRACT
\begin{abstract}
Myotonia, which refers to delayed muscle relaxation after contraction, is the main symptom of myotonic dystrophy patients. We propose a hierarchical attention-based temporal convolutional network (HA-TCN) for myotonic dystrohpy diagnosis from handgrip time series data, and introduce mechanisms that enable model explainability. We compare the performance of the HA-TCN model against that of benchmark TCN models, LSTM models with and without attention mechanisms, and SVM approaches with handcrafted features. In terms of classification accuracy and F1 score, we found all deep learning models have similar levels of performance, and they all outperform SVM. Further, the HA-TCN model outperforms its TCN counterpart with regards to computational efficiency regardless of network depth, and in terms of performance particularly when the number of hidden layers is small. Lastly, HA-TCN models can consistently identify relevant time series segments in the relaxation phase of the handgrip time series, and exhibit increased robustness to noise when compared to attention-based LSTM models. 
\end{abstract}

%%%%%%%%% BODY TEXT
\section{Introduction}\label{sec:intro}

According to a recent study~\cite{US44413318}, the volume of data in healthcare is projected to grow faster over the next seven years than in any of the other fields considered.  This is due to the increased frequency and resolution of data from multiple modalities being acquired in support of medical care. The rise of large datasets will likely be met with the development of artificial intelligence (AI) techniques aimed at distilling intelligence from so-called big data sources.  Ongoing research in fields including disease recurrence prediction \cite{giunchiglia2018rnn}, disease detection and classification \cite{hannun2019cardiologist}, and mortality prediction \cite{sha2017interpretable} offer a glance into the future landscape of AI-driven health analytics. Among candidate AI technologies, deep learning models are currently preferred as they avoid handcrafted features required by traditional machine learning approaches \cite{lin2018predicting, sha2017interpretable,zhou2016attention} and have shown promising results in the aforementioned tasks \cite{giunchiglia2018rnn, hannun2019cardiologist, sha2017interpretable}. 

In spite of their superior performance, deep learning models have been criticized for their lack of interpretability, particularly in the healthcare community \cite{wu2018beyond}. Model explainability is as important as accuracy in healthcare applications.  The certainty needs to exist that the diagnosis of a disease is made based on real causes instead of systemic biases in the data. Doctors and patients want to understand the reasoning process that leads to suggested treatment avenues. 

Recently, Wu \etal proposed tree regularization methodologies \cite{wu2018beyond} to enhance the explainability of deep learning models in high-dimensional medical time series classification tasks. Baumgartner \etal proposed a novel generative adversarial network (GAN)~\cite{maskGAN2017} to identify relevant visual features in medical image analysis tasks. In some cases, recurrent neural networks (RNNs) that leverage longitudinal and temporal dependencies in medical record data have been integrated with attention mechanisms to highlight relevant portions of electronic health records (EHR) \cite{sha2017interpretable, ma2017dipole}. Model-agnostic approaches that identify key predictors embedded in the data have also been proposed \cite{fong2017interpretable, lime2016}. 

In this study, we focus on interpretable diagnosis of myotonic dystrophy from analysis of handgrip time series data. Myotonic dystrophy is an autosomal dominant, progressive neuromuscular disorder caused by gene mutation \cite{heatwole2012patient, heatwole2015patient}. Its core feature, myotonia, consists in delayed muscle relaxation following contraction \cite{statland2012quantitative}, which heavily impacts a patient's daily life.  During a standardized quantitative myotonia assessment (QMA), a patient is required to squeeze a handgrip as hard as he/she can for 3 seconds, and value of the the maximal voluntary contraction (MVC) is recorded. The subject is then asked to release the grip as fast as possible. Multiple trials are usually conducted for each individual patient during each visit \cite{torres1983quantitative}. 

% data problem, challenges
In current automated analysis frameworks for myotonia diagnosis from QMA time series data, handcrafted features such as RT100 (i.e., the relaxation time it takes for the strength curve to fall from its maximum to baseline), RT90-5 (i.e., the relaxation time it takes the strength curve to fall from the 90\% to 5\% of its maximum value), etc.~are often used to differentiate between myotonic dystrophy patients and healthy subjects \cite{statland2012quantitative, torres1983quantitative}. However, the handgrip data can be noisy due to the high sensitivity of the QMA instrument, inconsistent data collection instructions, and variability in subjects' underlying medical conditions as well as their ability to follow the instructions. As a result, the extracted maximum strength value and baseline value estimates are inaccurate, and medical experts are often required to further verify their values and resulting diagnoses. %what else?

Our main contributions are as follows:\vspace{-0.3cm}
\begin{itemize}
    \item We propose an end-to-end hierarchical attention-based temporal convolutional network (HA-TCN) to automate the handgrip time series data analysis task.\vspace{-0.3cm}
    \item We empirically show that the proposed HA-TCN framework performs similarly to Temporal Convolutional Network (TCN) \cite{BaiTCN2018} and Long Short-term Memory (LSTM) approaches, and that the deep-learning-based methods outperform support vector machines (SVM) that rely on handcrafted features.\vspace{-0.3cm}
    \item We demonstrate through experimental validation that HA-TCN models outperform  TCNs in shallow architectural regimes because their hierarchical attention mechanisms enable them to better summarize relevant information across time steps.\vspace{-0.3cm}
    \item We empirically show that the HA-TCN model can highlight key time series segments in the relaxation phase of an individual handgrip sample that differentiate patients from healthy individuals. \vspace{-0.3cm}
\end{itemize}  %The HA-TCN is also more robust to the impact of noise and irrelevant signals when compared to the attention-based LSTMs on this task.  We point out that the latter two contributions are related to the superior explainability of our method. 

The remainder of this paper is organized as follows: Sec.~\ref{sec:related} overviews related work regarding attention mechanisms in deep learning models and their applications in the healthcare domain; Sec.~\ref{sec:model} delves into technical details regarding the proposed HA-TCN framework; Sec.~\ref{sec:exp} presents the experimental setup and results; lastly, Sec.~\ref{sec:concl} concludes the paper. 

%It is also important  in medical field. Most previous work are for image data. few has focused on time series data. Recently attention-based mechanism has been combined with recurrent neural network (RNN) for machine translation, dialogue act detection and so on. 

%------------------------------------------------------------------------
\section{Related Work}\label{sec:related}
\subsection{Attention Mechanisms}
Attention mechanisms were first proposed in \cite{bahdanau2014neural} for recurrent neural networks (RNN) on end-to-end machine translation applications. Since then, they have been widely applied to temporal sequence analysis. In an attention mechanism framework operating on an RNN, instead of stacking the information of the entire input sequence into one vector of fixed length \cite{tang2016sequence}, the sequence of latent vectors comprising the network activations is combined through a set of learned attention weights to form a context vector, which is then input to a feedforward layer for classification. % vinayavekhin2018focusing, tang2016sequence
 %The learned attention weights then indicate levels of focus on different parts in the temporal sequence. 
 The magnitude of the attention weights is proportional to the relevance of the corresponding segment to the classification decision. 

This kind of attention mechanism has been integrated with LSTMs in natural language processing (NLP). Shen and Lee~\cite{shen2016neural} applied an attention-mechanism LSTM for key term extraction and dialogue act detection. They showed that the attention mechanism enables the LSTM to ignore noisy or irrelevant passages in long input sequences, as well as to identify important portions of a given sequence to improve sequence labeling accuracy. % cosine similarity. some visualization results
Zhou \etal~\cite{zhou2016attention} proposed an attention-based bidirectional LSTM model for relation classification. %no visualization, learning weight 
Similar attention mechanisms have been implemented on top of deep learning models for time series classification and prediction tasks such as motion capture data completion and action classification \cite{vinayavekhin2018focusing, tang2016sequence}. 
%from tang2016sequence: Compared with other datasets such as images and voices, time series data is relatively sparse especially when one has in hand only a uni- channel temporal dataset. Crafting informative features from such a dataset demands imagination and is more of an art

Recently, the hierarchical attention network (HAN) model was proposed to select relevant encoder hidden states with two layers of attention mechanisms, namely relevant words and relevant sentences, for document classification \cite{yang2016hierarchical}. Inspired by these results, a dual-stage attention-based RNN for time series prediction was introduced for use in scenarios where multiple exogenous time series are available \cite{qin2017dual}. It was shown that the dual-stage attention mechanism can adaptively identify relevant exogenous time series at each time step, as well as salient encoder hidden states across all time steps.
% from yang2016hierarchical which summarizes the information of the whole sen- tence centered around wit. neighbor sentences around sentence i but still focus on sentence i.
\subsection{Attention-based Models for Healthcare}
Attention-based deep learning models have been applied for EHR data analysis. An EHR consists of longitudinal patient health data including demographics, diagnoses, and medications. The temporal, multimodal and high-dimensional nature of EHR data (comprising medical codes) makes it an ideal match for predictive modeling frameworks that leverage attention mechanisms with RNNs. Ma \etal implemented attention-based bidirectional RNNs for diagnosis prediction \cite{ma2017dipole}. Sha and Wang implemented a gated recurrent unit (GRU) network with hierarchical attention mechanisms for mortality prediction \cite{sha2017interpretable}. The hierarchical attention design can specify the importance of each hospital visit as well as the relevance of the different medical codes issued after each visit.   % learning weight limitation is it also only has 2~3 admissions for each patient 
%*********\cite{song2018attend}
In addition to the above decscribed applications in EHR data analysis, attention mechanisms have been applied in medical imaging studies, including image-based diagnosis of breast and colon cancer leveraging attention-based deep multi-instance learning \cite{ilse2018attention}, as well as brain hemorrhage detection from computed tomography (CT) scan analysis via recurrent attention densenet (RADnet) \cite{grewal2018radnet}. 

Even though attention mechanisms have a wide range of applications in healthcare data analysis, few studies have focused on classification of medical time series signals with attention-based deep learning models. Medical signals such as Intensive Care Unit (ICU) and electrocardiogram (ECG) data serve an important role in clinical decision-making. One potential reason for the scarcity of these types of models in the healthcare domain could be the lack of annotated data \cite{hannun2019cardiologist, lyu2018improving}. 
%\FloatBarrier
\begin{figure*}[h]
	\centering
	\includegraphics[width=15cm, height=5.5cm]{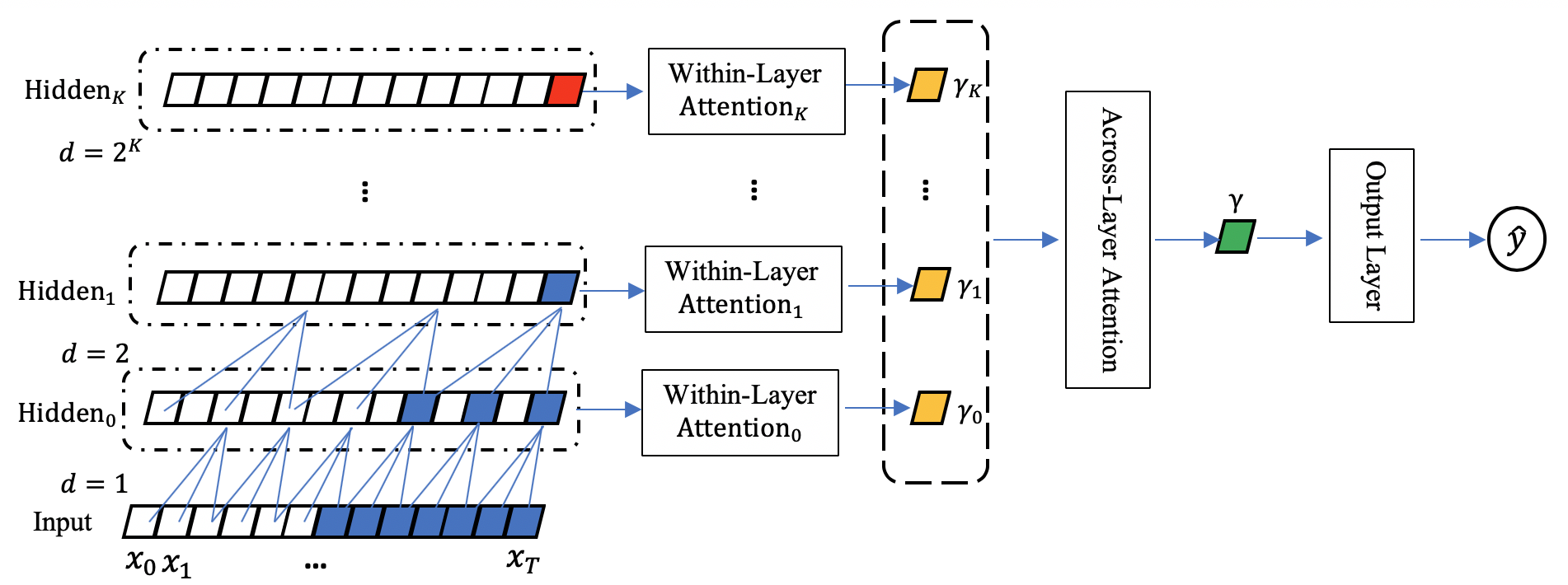}
	\caption{HA-TCN Architecture. This example network has one input layer and \(K\) hidden layers. The size of the kernel filter at each layer is 3. \(d\) is the dilation factor which increases exponentially with the layer number. The blue cells indicate the receptive field for the last cell in \(Hidden_1\) afforded by the dilated causal convolution mechanism. The red cell is the vector that the previous TCN \cite{BaiTCN2018} extracts for classification. The vectors \(\gamma_0, \gamma_1, ..., \gamma_K\) (in orange) represent the combination of convolutional activations through the within-layer attention layers, which are then converted into vector \(\gamma\) (in green) through the across-layer attention layer. \(\hat{y}\) is the classification result.}
\end{figure*}
%\FloatBarrier
\subsection{Temporal Convolutional Network}
Recently, it has been empirically demonstrated that TCNs consistently outperform RNNs such as LSTMs across a diverse range of benchmark time series modeling tasks \cite{BaiTCN2018}. TCNs have a few advantages over their RNN counterparts. Firstly, TCNs overcome the vanishing and exploding gradient issues associated with most RNNs due to the fact that their architecture is feedforward, and back propagation through time is not required during training. Secondly, the dilated convolutions in TCNs enable an exponentially increasing receptive field with network depth. In addition, TCNs are more highly parallelizable than RNNs, which makes their training and deployment more computationally feasible.

One aspect of TCNs that remains largely unexplored is explainability \cite{BaiTCN2018}. In this study, we incorporate a hierarchical attention mechanism on top of a TCN which results in a framework that is capable of accurate and interpretable diagnosis of myotonic dystrophy from handgrip time series data. The proposed model is able to identify segments of the input that are relevant to the automated decision-making by leveraging dilated convolutions and hierarchical attention mechanisms.

\section{HA-TCN Model}\label{sec:model}

Let \(X \in \mathbb{R} ^ T\), \(X = \{x_0, x_1, ..., x_T\}\) denote a time series sequence, e.g., a handgrip time series sample.  The decision-making task at hand consists in determining whether the subject whose series is being analyzed suffers from myotonic dystrophy.  Let \(y \in (0, 1) \) denote the output of the network, and \(f(\cdot)\) denote the input-output functional mapping effected by the network, i.e., \(y = f(X)\). The hierarchical attention mechanism outputs the relevant input portions which drive the classification decision. The architecture of HA-TCN is shown in \figurename{1}. 
%-------------------------------------------------------------------------
\subsection{Causal Convolutions and Dilated Convolutions}
The basic convolutional operations in the HA-TCN architecture are identical to those described in the original TCN publication \cite{BaiTCN2018}.  Specifically, the HA-TCN network uses a 1D fully-convolutional network (FCN) architecture and zero padding of length (kernel filter size - 1) to ensure inter-layer length compatibility. Causal convolutions and dilated convolutions are applied. The former ensures that only present (i.e., at time \(t\)) and past (i.e., before time \(t\)) samples are involved in the computation of the output at time \(t\). The latter introduces a fixed dilation factor \(d\) between every pair of adjacent filter taps. As previous studies indicate, \(d\) is increased exponentially with the depth of the network, e.g., \(d = O(2^i)\) for the \(i\)-th network layer. Assuming the kernel filter size is \(l\), its receptive field  (effective history) at its lower layer is \((l-1)d\), which also increases exponentially with layer depth. Note that we do not implement the residual connections used in \cite{BaiTCN2018}: only dilated causal convolutions are applied between two layers. 

%-------------------------------------------------------------------------
\subsection{Hierarchical Attention Mechanism}
When TCNs are used for classification tasks, the last sequential activation from the deepest layer is used (see red cell in \figurename{ 1}) \cite{TCNgithub}; as in RNNs, such activation condenses the information extracted from the entire input sequence into one vector.  This representation may be too abbreviated for complex sequential problems. With this in mind, we propose the addition of hierarchical attention mechanisms  across network layers. As shown in \figurename{ 1}, suppose the HA-TCN has \(K\) hidden layers, \(H_i\) is a matrix consisting of convolutional activations at layer \(i\), \(i=0, 1, ..., K\); \(H_i = [h_0^i, h_1^i, ..., h_T^i]\), \(H_i \in \Re ^ {C \times T}\), where \(C\) is the number of kernel filters at each layer. The within-layer attention weight \(\alpha_i \in \Re^{1 \times T}\) is calculated as follows:
\begin{equation}
\alpha_i = softmax(tanh(w_i^TH_i))
\end{equation}
where \(w_i \in \Re^{C \times 1}\) is a trained parameter vector, and \(w_i^T\) is its transpose. 

The combination of convolutional activations for layer \(i\) is calculated as:
\begin{equation}
\gamma_i = ReLU(H_i\alpha_i^T)
\end{equation}
where \(\gamma_i \in \Re^{C \times 1}\).

After executing each within-layer attention layer, the convolutional activations are now transformed into \(M = [\gamma_0, \gamma_1, ..., \gamma_i, ..., \gamma_K]\), \(M \in \Re ^ {C \times K}\). Similarly, the across-layer attention layer takes \(M\) as the input to calculate the final sequence representation used for classification:
\begin{equation}
\alpha = softmax(tanh(w^TM))
\end{equation}
\begin{equation}
\gamma = ReLU(M\alpha^T)
\end{equation}
where \(w \in \Re ^{C \times 1}\), \(\alpha \in \Re ^{1 \times K}\), \(\gamma \in \Re ^{C \times 1}\). 

%-------------------------------------------------------------------------
\subsection{Relevant Time Series Segment Identification}
In previous attention-based RNN studies, the relevant sequence segments corresponding to large attention weights were difficult to identify. This is due to the fact that the hidden state of a LSTM/RNN cell not only includes information from the current time step, but also includes information from historical time steps, the receptive field of which is unknown. For example, in the hierarchical attention network for document classification \cite{yang2016hierarchical}, the hidden state of a bidirectional GRU cell summarizes the information from the current sentence and also the neighboring sentences. However, the scope of the neighboring sentences is not clear.  

In contrast, the HA-TCN model can track the exact origin of relevant segments once the within-layer and across-layer attention estimates are available. This is because the architecture of the HA-TCN mainly consists of feedforward convolutional blocks. If \(d = 2^i\) in the dilated causal convolution operation, then the start of the receptive field at the input layer covered by a filter at time \(t\), layer \(i\) can be calculated as follows:
\begin{equation}
s = max(0, t - (2^{i+1} - 1)*(l-1))
%calculation process: max(0, t - ((2^0 + 2^1 + ... + 2^i)*(l - 1) + 1) + 1) = max(0, t - (2^{i+1} - 1)*(l-1))
\end{equation}
where \(s\) is the start time step of the receptive field at the input layer and \(l\) is the size of the kernel filter. 

The receptive field at the input layer for a filter at time step \(t\) and layer \(i\) can be then represented as \(RF_{s \rightarrow t}^i\), \(t=0, 1,..., T\), \(i = 0, 1, ...K\). As an example, in \figurename{ 1}, the receptive field for the filter at time step \(T\), hidden layer \(1\) is highlighted in blue. 

Given a series of handgrip data points, the within-layer attention \(\alpha_i\) and across-layer attention \(\alpha\) can be generated with a trained HA-TCN. We can rank hidden layers based on their \(\alpha\) value, and identify the relevant layers \(RL\) with larger attention weights (e.g., the top 10 percentile attention weights in \(\alpha\)). Similarly, we can also identify the relevant time steps \(RT\) with larger attention weights based on \(\alpha_i\), \(i \in RL\). 

Subsequently we can calculate the frequency of each time step \(j\) that belongs to the relevant field \(RF_{s \rightarrow t}^i\),  \(i \in RL\), and \(t \in RT\): 
\begin{equation}
Freq_j = \sum_{i \in RL, t \in RT}I_j(RF_{s \rightarrow t}^i), j = 0, 1, ..., T\
\end{equation}
where \(I_j(*)\) is the indicator function (i.e., valued 1 if \(j\) belongs to \(RF_{s \rightarrow t}^i\) and 0 otherwise). Lastly, the relevant time series segment can be identified based on corresponding high frequencies (e.g., the top 10 percentile frequencies). 

%\subsection{Entropy Regularization}
%entropy can be used to acquire sparse attention weights

\section{Experiments}\label{sec:exp}

%------------------------------------------------------------------------
\subsection{Dataset and Experimental Setup}

A total of 37 myotonic dystrophy patients and 18 healthy subjects were enrolled in this study.  According to the standard data collection process, each handgrip time series sample was collected by having the subject follow instructions to squeeze the QMA device (see \figurename{ 2(a)}), hold it with maximum force for about three seconds, and then release the grip. The subject's hand was held in place with pins so as to minimize the impact of any incidental arm movement on the acquired signal. Each trial consisted of six repetitions of this process, and each subject was asked to carry out two to three trials. Between squeezes within the same trial, the subject is allowed to rest for 10$\sim$12 seconds. The subject was allowed a 10 minute rest period Between trials. In total, 467 individual handgrip time series samples from patients and 270 samples from the healthy control group were acquired. 

The baseline performance benchmark comprises a classical machine learning framework leveraging a SVM operating on handcrafted features. In our case, handcrafted features are inspired by the way in which doctors perform diagnosis. A handgrip time series sample can be split into the squeezing phase, the maximum voluntary contraction (MVC) holding phase, and the relaxation phase. Because myotonia results in delayed muscle relaxation following contraction \cite{statland2012quantitative}, the relaxation phase contains the most discriminative features for classification. Previous studies associate the start of the relaxation phase with the instant when the MVC is attained \cite{torres1983quantitative}. However, factors like the sensitivity of the QMA device and severe myotonia symptoms can introduce noise in the acquired signal, which in turn makes it more difficult to determine the instant MVC takes place. We determine the start of the relaxation phase according to the following rules: (1) Define a strength threshold \(\eta = (\max(X) - \min(X)) / 2\), where ${\max(X)}$ and ${\min(X)}$ are the maximal and minimal strength, respectively; (2) Identify time steps \(TS\) with strength greater than \(\eta\); (3) Rank time steps in \(TS\) based on their strength and keep only points corresponding to the top 85\% strength values; (4) The start of the relaxation phase corresponds to the lattermost time step in the top 85\%. 

In this study, the relaxation time from the \(90^{th}\) percentile to the \(5^{th}\) percentile of the strength in the relaxation phase, i.e., RT90-5, is extracted to build traditional machine learning models. One example is shown in \figurename{ 2b}. The green cross indicates the start of the relaxation phase, and the time lapse between the red segment represents the extracted relaxation time for classification. 
%\FloatBarrier
\begin{figure}[h]
	\centering
	\begin{subfigure}[b]{0.35\textwidth}
		\includegraphics[width=1\linewidth]{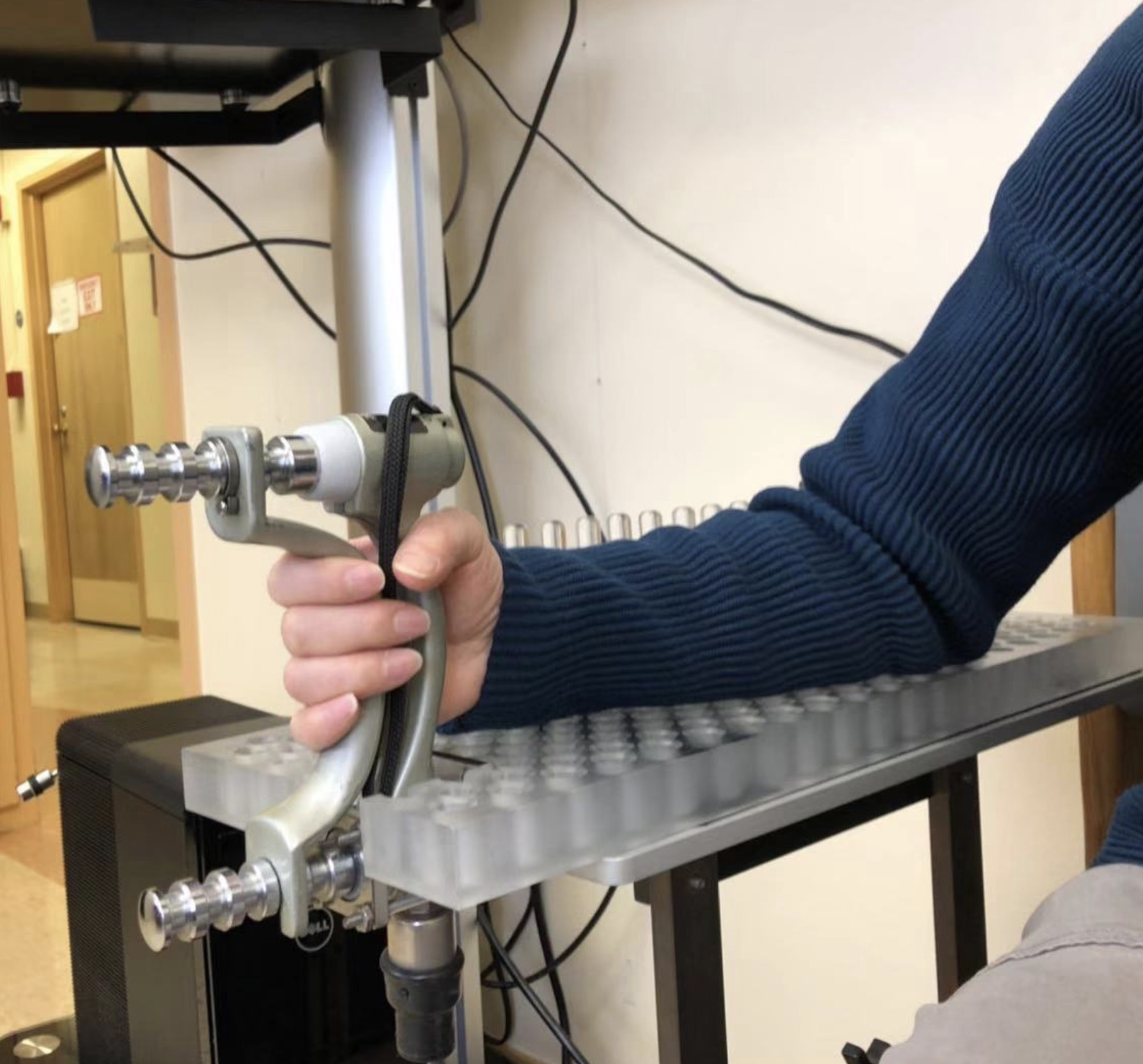}
		\caption{}
		\label{fig:Ng1} 
	\end{subfigure}
	
	\begin{subfigure}[b]{0.4\textwidth}
		\includegraphics[width=1\linewidth]{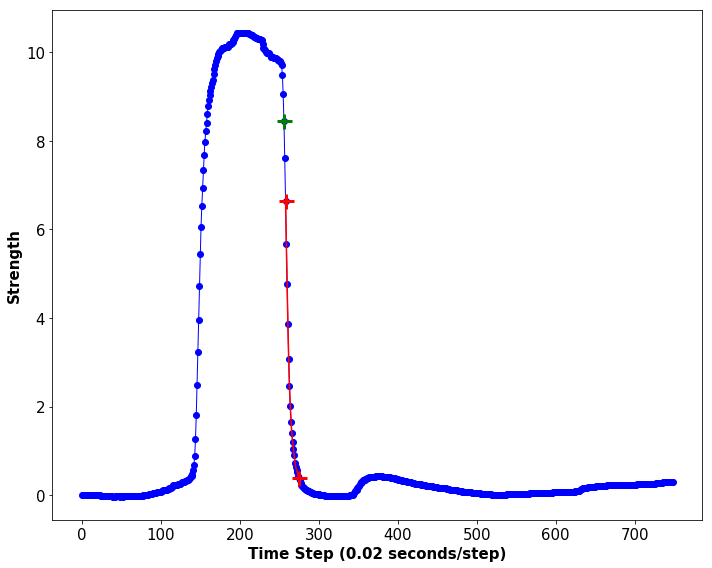}
		\caption{}
		\label{fig:Ng2}
	\end{subfigure}
	
	\caption[Two numerical solutions]{(a) Person squeezing the Quantitative Myotonia Assessment (QMA) device. (b) One handgrip time series sample. The green cross corresponds to the start of the relaxation phase; the red segment corresponds to the relaxation time from the $90^{th}$ percentile to the $5^{th}$ percentile of the strength in the relaxation stage.}
\end{figure}
%\FloatBarrier

We compare the performance of the proposed HA-TCN model to that of the classical SVM operating on the handcrafted features described above, and to those of recurrent architectures based on LSTMs and Bidirectional LSTMs (Bi-LSTM), and of recurrent architectures with attention mechanisms like attention-based LSTMs \cite{shen2016neural} and attention-based Bi-LSTMs \cite{zhou2016attention}, as well as that of the traditional TCN \cite{BaiTCN2018}. For deep learning models, each raw handgrip time series is truncated or padded to a total length of 750 time steps, and its amplitude normalized between \([0, 1]\). For LSTM models, the input length to each LSTM cell is set to 10, and one LSTM layer is applied. For the TCN model, we implement the architecture described in \cite{BaiTCN2018}, which consists of eight hidden layers of residual blocks, a kernel size of 7, and a dilation factor \(d\) of \(2^i\) for the \(i\)-th layer. The HA-TCN model consists of two hidden layers with dilated causal convolutions and dilation factors of 1 and 2, respectively, and a kernel size of 50. Ten-fold cross validation at the subject level is conducted, meaning that each subject is ensured to be included in the test set at least once. All the deep learning models are built in PyTorch. 

\begin{table}
	\begin{center}
		\begin{tabular}{|c|c|c|}
			\hline
			Model & Accuracy & F1 Score \\
			\hline\hline
			SVM & 88.40\% & 0.85 \\
			LSTM & $92.38\% \pm 2.49\%$& $0.94 \pm 0.01$  \\
			Bi-LSTM & $93.26\% \pm 1.85\%$ & $0.94 \pm 0.01$\\
			TCN \cite{BaiTCN2018} & $93.02\% \pm 2.38\%$ & $0.93 \pm 0.01$\\
			LSTM + Attention  & $93.58\% \pm 1.64\%$ & $0.94 \pm 0.01$\\
			Bi-LSTM + Attention \cite{zhou2016attention} & $ 94.00\% \pm 2.20\%$ & $0.94 \pm 0.01$\\
			HA-TCN & $93.82\% \pm 2.30\%$ & $0.95 \pm 0.01$ \\
			\hline
		\end{tabular}
	\end{center}
	\caption{Model Comparison in terms of Accuracy and F1 Score (mean \(\pm\) standard deviation). SVM performs the worst. Deep learning perform similarly. The attention mechanisms slightly improve model classification accuracy.}
\end{table}

%------------------------------------------------------------------------
\subsection{Experimental Results}
\subsubsection{Model Comparison}
%------------------------------------------------------------------------
The average classification accuracy and F1 score of each model are shown in \tablename{ 1}. For deep learning models, each fold is run five times to evaluate model consistency with respect to random weight initialization. The results show that the deep learning models outperform the SVM-based approach, which, as indicated, leverages handcrafted features resembling those used by doctors. All deep learning models have similar classification accuracies and F1 scores. Among those not leveraging attention mechanisms, Bi-LSTM has the best performance, while the TCN \cite{BaiTCN2018} model performs slightly better than the LSTM. The addition of attention mechanisms only slightly improves the performances of the corresponding deep learning models. This may be due to the nature of the dataset itself. 

Next, we compared HA-TCNs with TCNs \cite{BaiTCN2018}. A kernel size of 50 was used in both models, as well as the same dilation factor of \(2^i\), where \(i\) denotes the layer number. The main difference between the models is that the former implements dilated causal convolutions to generate hidden states and a hierarchical attention mechanism to combine information across time steps, while the latter utilizes more complex residual blocks for each layer (each block contains two layers of dilated causal convolutions and non-linearity transformations) and only uses the activation from the last cell of the deepest hidden layer for classification. Five runs of 10-fold cross validation were conducted with network depths ranging from 2 to 8 hidden layers. \figurename{ 3} shows the resulting average classification accuracy as well as the total run time for each architecture. 

The results in \figurename{ 3a} indicate that the proposed HA-TCN model reaches high classification accuracy with only two hidden layers, and that increasing network depth does not have a significant impact on model performance. In contrast, with two hidden layers, the classification accuracy of the TCN is relatively low. While its performance increases more significantly with network depth, it is lower than that of the HA-TCNs independently of network depth. We hypothesize this consistent difference in performance may be due to the reliance of the TCN on a single activation; this disadvantage can only be partially overcome with increasing network depth. \figurename{ 3b} also shows that the HA-TCN model always takes less time for training than the TCN models. 
%\FloatBarrier
\begin{figure}[h]
	\centering
	\begin{subfigure}[b]{0.35\textwidth}
		\includegraphics[width=1\linewidth, height=5cm]{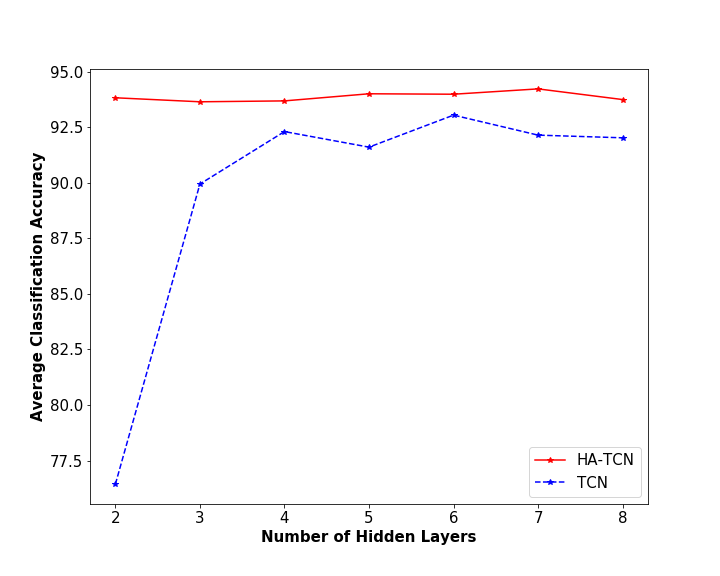}
		\caption{}
		\label{fig:Ng1} 
	\end{subfigure}
	
	\begin{subfigure}[b]{0.35\textwidth}
		\includegraphics[width=1\linewidth, height=5cm]{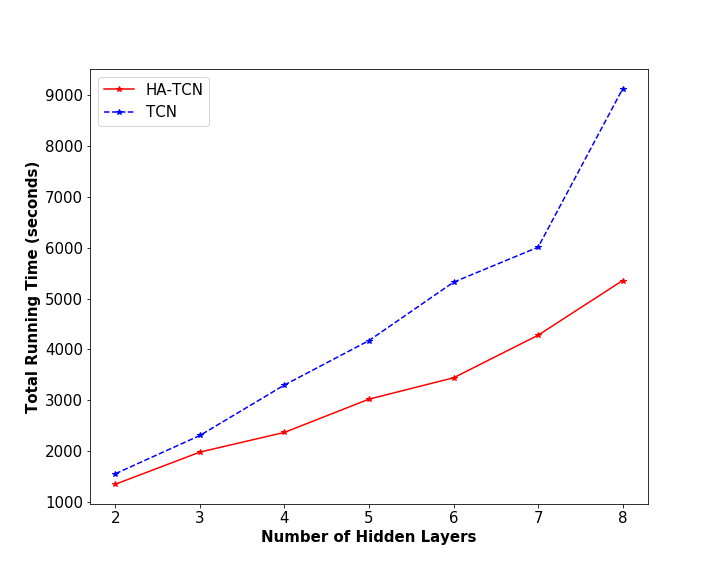}
		\caption{}
		\label{fig:Ng2}
	\end{subfigure}
	
	\caption{ Comparison of the HA-TCN and TCN models. (a) Average classification accuracy. (b) Total running time for five runs of 10-fold cross validation.}
\end{figure}
%\FloatBarrier
 
%  As mentioned in Sec.~\ref{sec:intro}, it is critical to ensure that the deep learning models are using relevant information, consistent with domain expert opinions, to make diagnostic classifications, i.e., the relaxation phase of a handgrip time series sample in this study. The next section will compare deep learning models from this aspect. 
 
 %------------------------------------------------------------------------
 \subsubsection{Key Time Series Segment Visualization}
\begin{figure*}[htb]
	\centering % <-- added
	\begin{subfigure}{0.33\textwidth}
		\includegraphics[width=\linewidth]{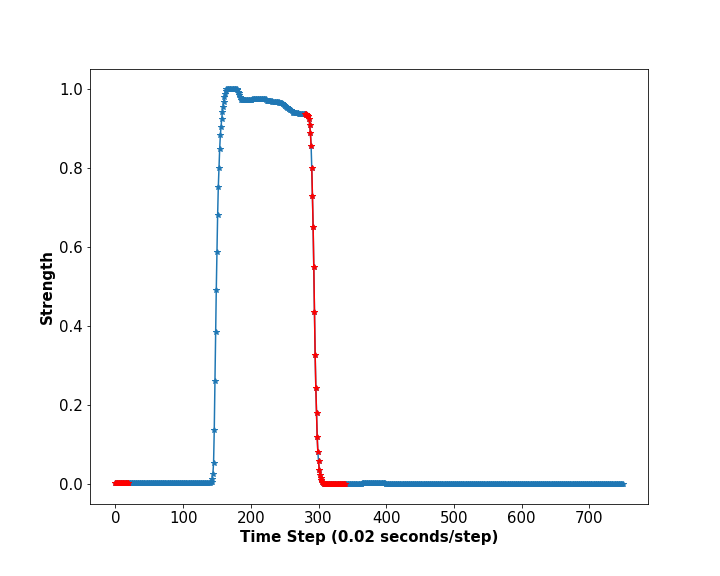}
		\caption{LSTM + Attention }
		\label{fig:1}
	\end{subfigure}\hfil % <-- added
	\begin{subfigure}{0.33\textwidth}
		\includegraphics[width=\linewidth]{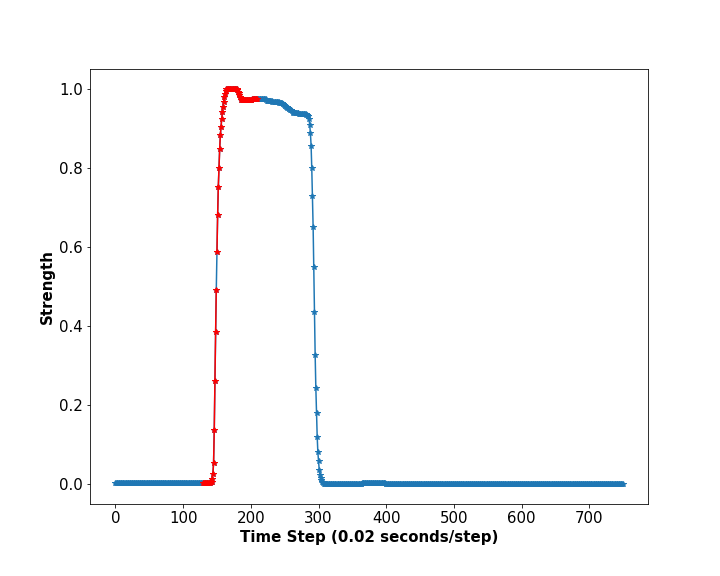}
		\caption{Bi-LSTM + Attention \cite{zhou2016attention}}
		\label{fig:2}
	\end{subfigure}\hfil % <-- added
	\begin{subfigure}{0.33\textwidth}
		\includegraphics[width=\linewidth]{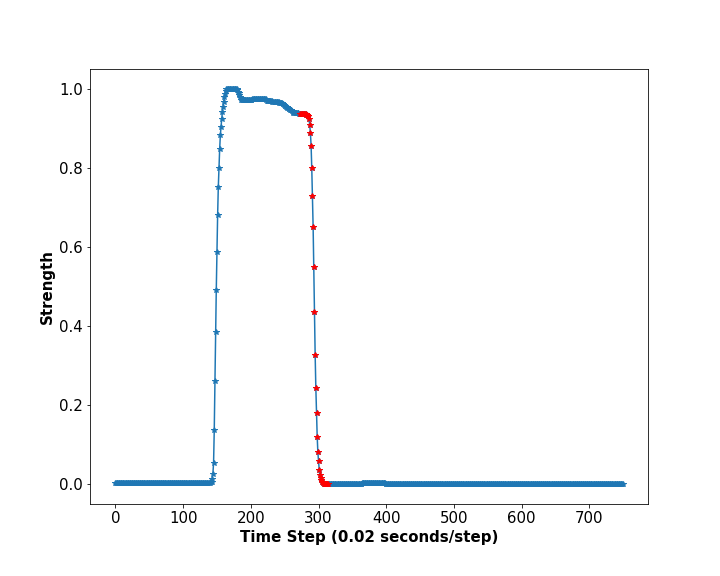}
		\caption{HA-TCN}
		\label{fig:3}
	\end{subfigure}
	
	\medskip
	\begin{subfigure}{0.33\textwidth}
		\includegraphics[width=\linewidth]{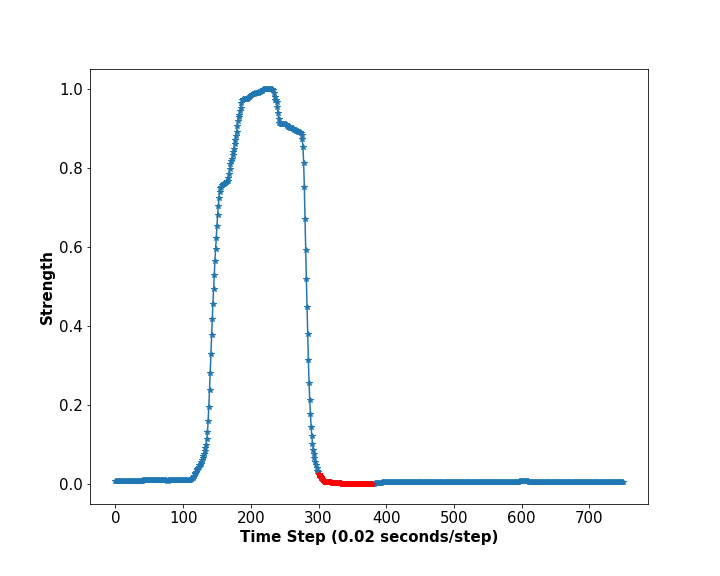}
		\caption{LSTM + Attention}
		\label{fig:4}
	\end{subfigure}\hfil % <-- added
	\begin{subfigure}{0.33\textwidth}
		\includegraphics[width=\linewidth]{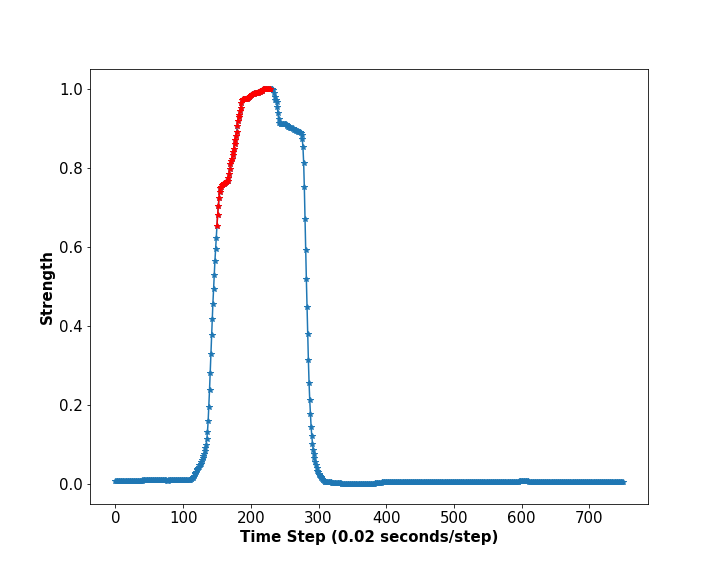}
		\caption{Bi-LSTM + Attention \cite{zhou2016attention}}
		\label{fig:5}
	\end{subfigure}\hfil % <-- added
	\begin{subfigure}{0.33\textwidth}
		\includegraphics[width=\linewidth]{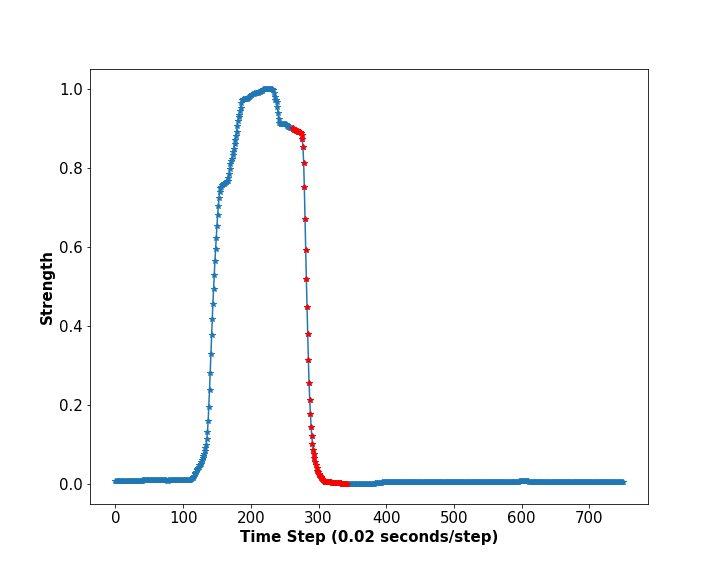}
		\caption{HA-TCN}
		\label{fig:6}
	\end{subfigure}
	\caption{Relevant segment identification with attention-based deep learning models. (a), (b), (c) are  handgrip time series samples from a healthy subject; (d), (e), (f) are handgrip time series samples from a patient. Red segments are identified as relevant segments from the deep learning models.}
	\label{fig:images}
\end{figure*}

 %------------------------------------------------------------------------
 \figurename{ 4a} and \figurename{ 4b} show the average attention weight for each time step for the LSTM models. \figurename{ 4c} shows the average frequency that a time step belongs to a specific receptive field. Because the HA-TCN model in this study only has two hidden layers, we select the one with the larger across-layer attention weight as the relevant layer \(RL\), then choose those with the top 10 percentile attention weights as the relevant time steps \(RT\). The frequency of a time step belonging to the receptive fields is calculated based on Equations (5) and (6). In addition to the average weight and frequency curves for all subjects, the plots in \figurename{ 4} also show the corresponding curves for patients and healthy subjects separately. 
 
 In \figurename{ 4a}, the maximum average attention weight for patients occurs at around time step 400. For healthy patients, that maximum occurs at time step 310. Similarly, in \figurename{ 4b}, the peak of the average attention weight curve for the healthy group is again at around the 300\(^{th}\) time step. This curve for the patients is, however, flatter, and the time steps around the 200\(^{th}\) time step all have large attention weights. The average attention weight curve based on all subjects shows that the segment between the 200\(^{th}\) to the 300\(^{th}\) time step is more relevant to the decision-making process. These inconsistencies indicate that the attention-based LSTMs cannot be used for interpretable diagnosis of myotonic dystrophy. In contrast, it can be seen in \figurename{ 4c} that the curves for the HA-TCN model across subjects, including patients and healthy subjects, almost align perfectly, with peaks taking place at around the 350 \(^{th}\) time step. 
 
%\FloatBarrier
\begin{figure*}
	\begin{subfigure}{.33\textwidth}
		\centering
		\includegraphics[width=6cm,height=6cm]{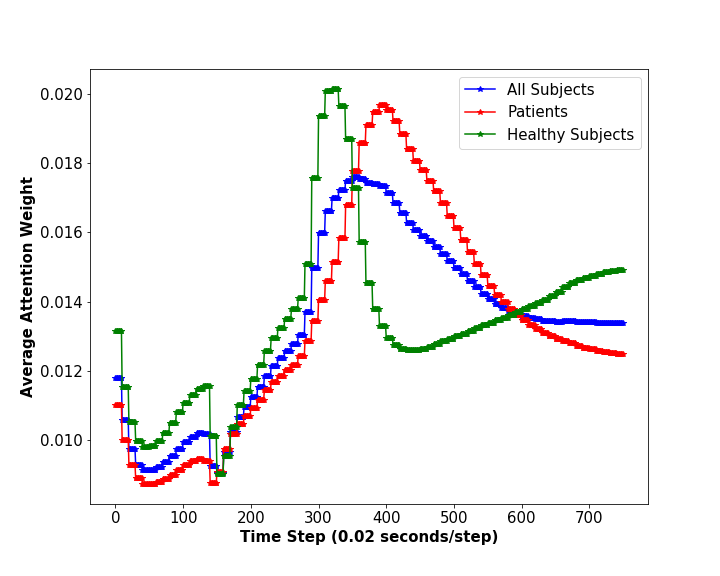}
		\caption{LSTM + Attention}
	\end{subfigure}\hfill
	\begin{subfigure}{.33\textwidth}
		\centering
		\includegraphics[width=6cm,height=6cm]{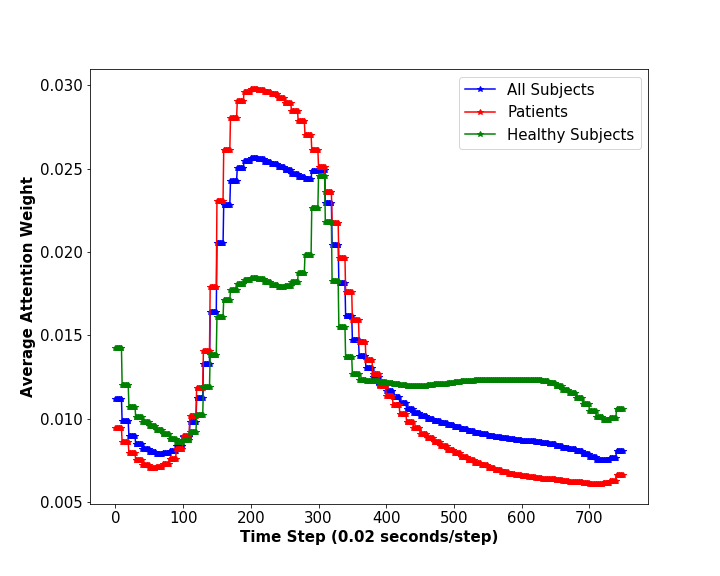}
		\caption{Bi-LSTM + Attention \cite{zhou2016attention}}
	\end{subfigure}\hfill
	\begin{subfigure}{.33\textwidth}
		\centering
		\includegraphics[width=6cm,height=6cm]{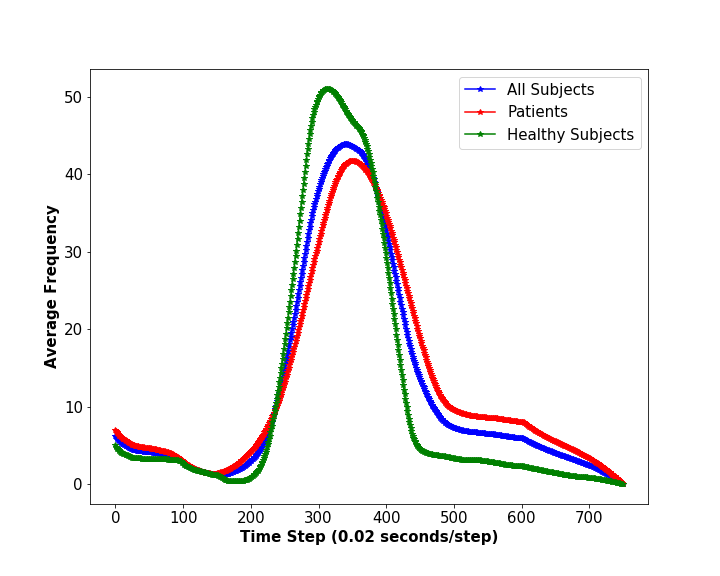}
		\caption{HA-TCN}
	\end{subfigure}\hfill
	\caption{Explainability Comparison of Deep learning models. (a) and (b) are average attention weight per time step for LSTMs (the attention weight is the same for the 10 time steps in one LSTM cell; (c) is the average frequency that the time step is within the receptive fields based on the HA-TCN. }
\end{figure*}
%\FloatBarrier

 \figurename{ 5} shows relevant time series segments identified by attention-based deep learning models for two examples. The time series data in (a), (b) and (c) belongs to a healthy subject, and the data in (d), (e), and (f) is sampled from a patient. The red segments identified by the LSTM models are the time steps with the top 10 percentile attention weights, and the red segments found by the HA-TCN model in \figurename{ 5c} and \figurename{ 5d} are those with the top 10 percentile frequencies. It can be seen that one-directional LSTM models with attention highlight the strength decreasing part for the healthy subject in \figurename{ 5a}, but the beginning of the curve is also identified as important. For the patient time series curve in \figurename{ 5d}, the most relevant segment is located at the end of the relaxation phase, which is inconsistent with the definition of myotonia. For red segments identified by the Bi-LSTM model with attention in \figurename{ 5b} and 5e, the segments corresponding to the squeezing stage are assigned high weights, which again shows that such model is not usable for interpretable diagnosis of myotonic dystrophy. As shown in \figurename{ 5c} and 5f, only the HA-TCN model can consistently identify the deceasing portion of the time series as the part most relevant for disease classification. 

%\FloatBarrier
%\FloatBarrier

\section{Conclusion}\label{sec:concl}
In this paper, we introduced the hierarchical attention-based temporal convolutional network (HA-TCN) architecture for interpretable diagnosis of myotonic dystrophy patients. The HA-TCN model achieves performances comparable with state-of-the-art deep learning models. All deep learning models outperform traditional machine learning approaches relying on handcrafted features. However, only the HA-TCN can highlight the relaxation phase in the handgrip time series data, which is consistent with the diagnosis criterion that clinicians have been using. Furthermore, we show that the HA-TCN outperforms the TCN in terms of performance particularly when the number of hidden layers is small and it is more computationally efficient in general. 

For future research directions, we will test the performances of the HA-TCNs on multiple datasets. It is also interesting to compare the explainability of the HA-TCNs with previous works that utilize other approaches instead of attention mechanisms such as \cite{maskGAN2017, fong2017interpretable, lime2016}. 

{\small
\bibliographystyle{ieee}
\bibliography{egbib}

\begin{thebibliography}{10}\itemsep=-1pt

\bibitem{TCNgithub}
Sequential mnist and permuted sequential mnist.
\newblock \url{https://github.com/locuslab/TCN/tree/master/TCN/mnist_pixel}.
\newblock Accessed: 2019-03-07.

\bibitem{bahdanau2014neural}
D.~Bahdanau, K.~Cho, and Y.~Bengio.
\newblock Neural machine translation by jointly learning to align and
  translate.
\newblock {\em arXiv preprint arXiv:1409.0473}, 2014.

\bibitem{BaiTCN2018}
S.~Bai, J.~Z. Kolter, and V.~Koltun.
\newblock An empirical evaluation of generic convolutional and recurrent
  networks for sequence modeling.
\newblock {\em arXiv:1803.01271}, 2018.

\bibitem{maskGAN2017}
C.~F. Baumgartner, L.~M. Koch, K.~C. Tezcan, J.~X. Ang, and E.~Konukoglu.
\newblock Visual feature attribution using wasserstein gans.
\newblock {\em CoRR}, abs/1711.08998, 2017.

\bibitem{fong2017interpretable}
R.~C. Fong and A.~Vedaldi.
\newblock Interpretable explanations of black boxes by meaningful perturbation.
\newblock In {\em Proceedings of the IEEE International Conference on Computer
  Vision}, pages 3429--3437, 2017.

\bibitem{giunchiglia2018rnn}
E.~Giunchiglia, A.~Nemchenko, and M.~van~der Schaar.
\newblock Rnn-surv: A deep recurrent model for survival analysis.
\newblock In {\em International Conference on Artificial Neural Networks},
  pages 23--32. Springer, 2018.

\bibitem{grewal2018radnet}
M.~Grewal, M.~M. Srivastava, P.~Kumar, and S.~Varadarajan.
\newblock Radnet: Radiologist level accuracy using deep learning for hemorrhage
  detection in ct scans.
\newblock In {\em 2018 IEEE 15th International Symposium on Biomedical Imaging
  (ISBI 2018)}, pages 281--284. IEEE, 2018.

\bibitem{hannun2019cardiologist}
A.~Y. Hannun, P.~Rajpurkar, M.~Haghpanahi, G.~H. Tison, C.~Bourn, M.~P.
  Turakhia, and A.~Y. Ng.
\newblock Cardiologist-level arrhythmia detection and classification in
  ambulatory electrocardiograms using a deep neural network.
\newblock {\em Nature medicine}, 25(1):65, 2019.

\bibitem{heatwole2012patient}
C.~Heatwole, R.~Bode, N.~Johnson, C.~Quinn, W.~Martens, M.~P. McDermott,
  N.~Rothrock, C.~Thornton, B.~Vickrey, D.~Victorson, et~al.
\newblock Patient-reported impact of symptoms in myotonic dystrophy type 1
  (prism-1).
\newblock {\em Neurology}, 79(4):348--357, 2012.

\bibitem{heatwole2015patient}
C.~Heatwole, N.~Johnson, R.~Bode, J.~Dekdebrun, N.~Dilek, J.~E. Hilbert,
  E.~Luebbe, W.~Martens, M.~P. McDermott, C.~Quinn, et~al.
\newblock Patient-reported impact of symptoms in myotonic dystrophy type 2
  (prism-2).
\newblock {\em Neurology}, 85(24):2136--2146, 2015.

\bibitem{ilse2018attention}
M.~Ilse, J.~M. Tomczak, and M.~Welling.
\newblock Attention-based deep multiple instance learning.
\newblock {\em arXiv preprint arXiv:1802.04712}, 2018.

\bibitem{lin2018predicting}
L.~Lin, Z.~He, and S.~Peeta.
\newblock Predicting station-level hourly demand in a large-scale bike-sharing
  network: A graph convolutional neural network approach.
\newblock {\em Transportation Research Part C: Emerging Technologies},
  97:258--276, 2018.

\bibitem{lyu2018improving}
X.~Lyu, M.~Hueser, S.~L. Hyland, G.~Zerveas, and G.~Raetsch.
\newblock Improving clinical predictions through unsupervised time series
  representation learning.
\newblock {\em arXiv preprint arXiv:1812.00490}, 2018.

\bibitem{ma2017dipole}
F.~Ma, R.~Chitta, J.~Zhou, Q.~You, T.~Sun, and J.~Gao.
\newblock Dipole: Diagnosis prediction in healthcare via attention-based
  bidirectional recurrent neural networks.
\newblock In {\em Proceedings of the 23rd ACM SIGKDD international conference
  on knowledge discovery and data mining}, pages 1903--1911. ACM, 2017.

\bibitem{qin2017dual}
Y.~Qin, D.~Song, H.~Chen, W.~Cheng, G.~Jiang, and G.~Cottrell.
\newblock A dual-stage attention-based recurrent neural network for time series
  prediction.
\newblock {\em arXiv preprint arXiv:1704.02971}, 2017.

\bibitem{US44413318}
D.~Reinsel, J.~Gantz, and J.~Rydning.
\newblock White paper: The digitization of the world from edge to core.
\newblock Technical Report US44413318, IDC, November 2018.

\bibitem{lime2016}
M.~T. Ribeiro, S.~Singh, and C.~Guestrin.
\newblock "why should {I} trust you?": Explaining the predictions of any
  classifier.
\newblock {\em CoRR}, abs/1602.04938, 2016.

\bibitem{sha2017interpretable}
Y.~Sha and M.~D. Wang.
\newblock Interpretable predictions of clinical outcomes with an
  attention-based recurrent neural network.
\newblock In {\em Proceedings of the 8th ACM International Conference on
  Bioinformatics, Computational Biology, and Health Informatics}, pages
  233--240. ACM, 2017.

\bibitem{shen2016neural}
S.-s. Shen and H.-y. Lee.
\newblock Neural attention models for sequence classification: Analysis and
  application to key term extraction and dialogue act detection.
\newblock {\em arXiv preprint arXiv:1604.00077}, 2016.

\bibitem{statland2012quantitative}
J.~M. Statland, B.~N. Bundy, Y.~Wang, J.~R. Trivedi, D.~Raja~Rayan,
  L.~Herbelin, M.~Donlan, R.~McLin, K.~J. Eichinger, K.~Findlater, et~al.
\newblock A quantitative measure of handgrip myotonia in non-dystrophic
  myotonia.
\newblock {\em Muscle \& nerve}, 46(4):482--489, 2012.

\bibitem{tang2016sequence}
Y.~Tang, J.~Xu, K.~Matsumoto, and C.~Ono.
\newblock Sequence-to-sequence model with attention for time series
  classification.
\newblock In {\em 2016 IEEE 16th International Conference on Data Mining
  Workshops (ICDMW)}, pages 503--510. IEEE, 2016.

\bibitem{torres1983quantitative}
C.~Torres, R.~T. Moxley, and R.~C. Griggs.
\newblock Quantitative testing of handgrip strength, myotonia, and fatigue in
  myotonic dystrophy.
\newblock {\em Journal of the neurological sciences}, 60(1):157--168, 1983.

\bibitem{vinayavekhin2018focusing}
P.~Vinayavekhin, S.~Chaudhury, A.~Munawar, D.~J. Agravante, G.~De~Magistris,
  D.~Kimura, and R.~Tachibana.
\newblock Focusing on what is relevant: Time-series learning and understanding
  using attention.
\newblock In {\em 2018 24th International Conference on Pattern Recognition
  (ICPR)}, pages 2624--2629. IEEE, 2018.

\bibitem{wu2018beyond}
M.~Wu, M.~C. Hughes, S.~Parbhoo, M.~Zazzi, V.~Roth, and F.~Doshi-Velez.
\newblock Beyond sparsity: Tree regularization of deep models for
  interpretability.
\newblock In {\em Thirty-Second AAAI Conference on Artificial Intelligence},
  2018.

\bibitem{yang2016hierarchical}
Z.~Yang, D.~Yang, C.~Dyer, X.~He, A.~Smola, and E.~Hovy.
\newblock Hierarchical attention networks for document classification.
\newblock In {\em Proceedings of the 2016 Conference of the North American
  Chapter of the Association for Computational Linguistics: Human Language
  Technologies}, pages 1480--1489, 2016.

\bibitem{zhou2016attention}
P.~Zhou, W.~Shi, J.~Tian, Z.~Qi, B.~Li, H.~Hao, and B.~Xu.
\newblock Attention-based bidirectional long short-term memory networks for
  relation classification.
\newblock In {\em Proceedings of the 54th Annual Meeting of the Association for
  Computational Linguistics (Volume 2: Short Papers)}, volume~2, pages
  207--212, 2016.

\end{thebibliography}
}

\end{document}